\documentclass[10pt, a4paper]{article}
\usepackage[T1]{fontenc}
\usepackage{lrec2022} % this is the new LREC2022 Style
\usepackage{multibib}
\newcites{languageresource}{Language Resources}
\usepackage{graphicx}
\usepackage{tabularx}
\usepackage{soul}
\usepackage{titlesec}
\titleformat{\section}{\normalfont\large\bfseries\center}{\thesection.}{1em}{}
\titleformat{\subsection}{\normalfont\SmallTitleFont\bfseries\raggedright}{\thesubsection.}{1em}{}
\titleformat{\subsubsection}{\normalfont\normalsize\bfseries\raggedright}{\thesubsubsection.}{1em}{}
\renewcommand\thesection{\arabic{section}}
\renewcommand\thesubsection{\thesection.\arabic{subsection}}
\renewcommand\thesubsubsection{\thesubsection.\arabic{subsubsection}}
\usepackage{epstopdf}
\usepackage[utf8]{inputenc}

\usepackage{xstring}

\usepackage{color}
\usepackage{booktabs}

\usepackage[title]{appendix}

\makeatletter
\newcommand{\printfnsymbol}[1]{%
  \textsuperscript{\@fnsymbol{#1}}%
}
\makeatother
\usepackage{hyperref}

\title{A Warm Start and a Clean Crawled Corpus -\\ A Recipe for Good Language Models}

\name{Vésteinn Snæbjarnarson$^{1,2,}$\printfnsymbol{1}\begin{NoHyper}\thanks{\printfnsymbol{1} Equal contribution.}\end{NoHyper}, Haukur Barri Símonarson$^{1,2,}$\printfnsymbol{1}, \\ {\bf \large Pétur Orri Ragnarsson$^1$, Svanhvít Lilja Ingólfsdóttir$^1$, Haukur Páll Jónsson$^1$,}\\ {\bf \large Vilhjálmur Þorsteinsson$^1$, Hafsteinn Einarsson$^2$}}

\address{$^1$ Miðeind ehf.,  $^2$ University of Iceland\\
         \{vesteinn, haukur, petur, svanhvit, haukurpj, vt\}@mideind.is, hafsteinne@hi.is\\}

\abstract{
We train several language models for Icelandic, including IceBERT, that achieve state-of-the-art performance in a variety of downstream tasks, including part-of-speech tagging, named entity recognition, grammatical error detection and constituency parsing. To train the models we introduce a new corpus of Icelandic text, the Icelandic Common Crawl Corpus (IC3), a collection of high quality texts found online by targeting the Icelandic top-level-domain \texttt{.is}. Several other public data sources are also collected for a total of 16GB of Icelandic text. To enhance the evaluation of model performance and to raise the bar in baselines for Icelandic, we manually translate and adapt the WinoGrande commonsense reasoning dataset. Through these efforts we demonstrate that a properly cleaned crawled corpus is sufficient to achieve state-of-the-art results in NLP applications for low to medium resource languages, by comparison with models trained on a curated corpus. We further show that initializing models using existing multilingual models can lead to state-of-the-art results for some downstream tasks. \\ \newline  }
\begin{document}

\maketitleabstract

\section{Introduction}
The Government of Iceland recently launched an initiative to improve the state of Icelandic language resources and language technology~\cite{nikulasdottir_language_2020}. This comprehensive program has its roots in the historical focus on protecting the Icelandic language~\cite{kristinsson2018national}, and, as a result, work has been ongoing to build and enhance said resources. This effort is gradually pushing Icelandic from being a low-resource language to a medium-resource one\footnote{\textit{Low-resource} is not a precise term, but a language can be considered to be low-resource if few online resources exist for it \cite{cieri-etal-2016-selection}.}. Still, and apart from large monolingual corpora~\citelanguageresource{risamalheild}, many important types of resources are lacking in comparison with major languages such as English.

Parallel to the development of the Icelandic language technology program, language technology world-wide has been progressing at a fast and accelerating pace \cite{foundationmodel}. Pre-trained neural language models based on Transformers~\cite{attentionvaswani} have shown impressive results when adapted for a variety of classification and text generation tasks. Such models are now applied widely across industries and modalities.

%Large language models such as BERT~\cite{bertdevlin}, BART~\cite{lewis_bart_2019} and GPT-2~\cite{radford2019language} have historically first been trained and evaluated on English, or on a collection of languages where English is dominant, as English is the language with the most developed and most extensive language resources. %As a result, it is the default language when comparing a given model's performance with other models. English is also widely spoken and, thus, economically interesting for research divisions within large technology companies, where much of the work in natural language processing takes place.
Large monolingual language models such as BERT~\cite{bertdevlin}, BART~\cite{lewis_bart_2019} and GPT-2~\cite{radford2019language} have been developed for English. For many smaller languages the only available options are multilingual models, which can reach impressive performance on downstream tasks~\cite{conneau_unsupervised_2020}. These are not without their flaws though; when compared to training on a sufficiently large monolingual corpus in a given language, multilingual models can lead to less than optimal performance, as demonstrated in the case of Finnish~\cite{virtanen_multilingual_2019}. Since an evaluation of Transformer language models for Icelandic is yet to be completed, it has remained unclear to what extent this holds for Icelandic.

The data used to train language models is usually sourced from large collections of books (e.g. \citelanguageresource{Zhu_2015_ICCV}) and online texts, where the choice and quality of training data can potentially have a large effect on downstream task performance. While curated corpora may not be readily available for a language, it might still be relatively well represented online, in the form of web texts, which can be sourced by automatic means. This raises the question of whether language models trained on curated corpora offer better performance in downstream tasks than those trained predominantly on data sourced from the web.

In this paper, we show how a Transformer model, IceBERT, can be trained for Icelandic with relatively modest language resources to reach state-of-the-art performance across a variety of tasks. We train multiple models on monolingual corpora from different sources: a curated corpus (Icelandic Gigaword Corpus, IGC \citelanguageresource{risamalheild}) and a corpus of text collected efficiently from Common Crawl\footnote{\href{https://commoncrawl.org/}{\texttt{commoncrawl.org}}}. 

We train separate models on each of these two sources, and compare results, to demonstrate the feasibility of our approach for other languages with similar resource availability. To our surprise, models trained on texts extracted from the web achieved similar performance to models trained on a curated corpus. We also evaluate the performance of a multilingual model (XLMR-base, \newcite{conneau_unsupervised_2020}), which shows good results for some tasks but is insufficient for others. Finally, we use the existing multilingual model as a warm start and continue pre-training on Icelandic text, reaching state-of-the-art results in downstream tasks such as NER.

While large corpora of multilingual text exist, such as the multilingual Colossal Clean Crawled Corpus (mC4)~\cite{xue_mt5_2021} that is sourced from the Common Crawl, they have not been officially released in a way such that text in smaller languages can be easily extracted. However, the mC4 dataset has been made available by a third party\footnote{\href{https://huggingface.co/datasets/mc4}{\texttt{huggingface.co/datasets/mc4}}} which we use for our experiments. Additionally, we demonstrate how to directly extract Icelandic text from the Common Crawl in a novel way, explain how it can be done for other languages, and highlight the importance of clean data.

Regarding applicability of our approach to other languages, we note that in mC4, there are 107 labelled languages, with almost half of the 6.7 billion documents being in English. The average number of documents in languages other than English is 33.4 million documents per language, and the median value is 2 million documents, which happens to be the approximate number for Icelandic \cite{xue_mt5_2021}. 
These numbers show that our approach of extracting training data should be well within reach of at least half of the languages in Common Crawl, and possibly applicable for the 46 languages containing between 500 thousand and 10 million documents.

\textbf{The key contributions of our work} are summarized below.

\textbf{(a) Several Icelandic language models}, including IceBERT,\footnote{Available at \href{https://huggingface.co/mideind/}{\texttt{huggingface.co/mideind}}.} trained on a monolingual corpus with 2.7B tokens.

\textbf{(b) Adaptations of IceBERT} with state-of-the-art results for part-of-speech tagging (PoS), named entity recognition (NER), constituency parsing and grammatical error detection (GED).

\textbf{(c) The Icelandic Common Crawl Corpus (IC3)}\footnote{We have made the dataset available at \url{https://huggingface.co/datasets/mideind/icelandic-common-crawl-corpus-IC3}}, a cleaned and deduplicated corpus extracted by targeting the \texttt{.is} top level domain.

\textbf{(d) The Icelandic WinoGrande dataset (IWG)}\footnote{Will be made available on the Icelandic CLARIN repository \href{https://repository.clarin.is}{\texttt{repository.clarin.is}}.}, a new and challenging benchmark for commonsense reasoning and natural language understanding.

\section{Related work}
The original BERT model has, since its publishing, spawned a whole family of BERT-like models. One of the main reasons for their popularity is their potential for transfer learning, i.e.\ the possibility to adapt them and obtain impressive performance on benchmarks and tasks that they were not originally trained for.

Multilingual versions of BERT exist that are trained on text in multiple languages, such as mBERT, which is trained on Wikipedia in over a hundred languages including Icelandic. Since the release of BERT, other large pre-trained models such as mBART~\cite{liu-etal-2020-multilingual-denoising} and XLMR~\cite{conneau_cross-lingual_2019,conneau_unsupervised_2020} have been trained that include Icelandic and other lower-resource languages. In addition, mT5~\cite{xue_mt5_2021}, a sequence-to-sequence model, is trained on the entire mC4 corpus.

Multilingual models are often the only option for low-resource languages, which do not have direct access to sufficient language data or computational resources to create monolingual transformer-based language models. Such multilingual models have been shown to have useful properties, including zero-shot crosslingual transfer. That is, fine-tuning these models on a downstream task in one language can translate to improved performance in other languages without explicit crosslingual signals~\cite{pires_how_2019,wu_beto_2019,k_cross-lingual_2020}.

%It remains unclear what is the ideal setting to pre-train a model to reach good results for a given language.
Despite the impressive results for multilingual models, they might not be the right choice where output accuracy is critical. For some languages, it may be better to pre-train a model on a monolingual corpus and adapt it for downstream tasks rather than to adapt a model trained on a multilingual corpus, as in the case of Finnish~\cite{virtanen_multilingual_2019}. It has also been shown that the crosslingual capabilities of mBERT only apply to high-resource languages~\cite{wu_are_2020}. Furthermore, benchmarks of crosslingual transfer show a sizable gap in the performance of crosslingually transferred models when compared to monolingually trained ones~\cite{hu_xtreme_2020}. These results highlight the still basic need for more training data in the case of medium- and low-resource languages.

% Tafla úr mT5 grein: https://www.semanticscholar.org/paper/mT5%3A-A-Massively-Multilingual-Pre-trained-Xue-Constant/74276a37bfa50f90dfae37f767b2b67784bd402a/figure/5
\begin{table*}[]
\begin{tabular*}{\textwidth}{@{}llll@{}}
\toprule
\textbf{Language} & \textbf{mC4 tokens (B)} & \textbf{Native speakers} & \textbf{Reference} \\ \midrule
English   & 2,733  & 380M &  \cite{bertdevlin}\\
%French    & 318 & 76.8M & \cite{martin_camembert_2020,le_flaubert_2020}\\
%Italian   & 162 & 67M & \cite{polignano_alberto_2019}\\
%Dutch     & 73 & 25M & \cite{delobelle_robbert_2020,de_vries_bertje_2019}\\
%Hungarian & 39 & 13M & \cite{Nemeskey:2020}\\
%Finnish   & 25 & 5.8M & \cite{virtanen_multilingual_2019}\\
Icelandic & 2.6 & 350k & (This paper) \\
Galician & 2.4 & 2.4M & \cite{vilares_bertinho_2021}\\
Urdu (Roman) & 2.4 & 70M & \cite{khalid_rubert_2021}\\
Filipino &  2.1  & 23.8M & \cite{cruz_establishing_2020}\\
Afrikaans &  1.7  & 7.2M & \cite{ralethe_adaptation_2020}\\ % Multi yfir í mono
Basque & 1.4  & 900k & \cite{agerri_give_2020}\\
Telugu & 1.3 & 83M & \cite{marreddy_clickbait_2021}\\
Latin & 1.3 & 0 & \cite{bamman_latin_2020}\\
Swahili & 1.0 & 18M & \cite{bhattacharjee_banglabert_2021}\\\bottomrule
\end{tabular*}
\caption{Language models trained on a monolingual corpus. We highlight some languages that have a similar number of tokens to Icelandic in mC4. The number of speakers denotes the number of native speakers (L1) according to the Wikipedia page for each language.}
\label{tab:1}
\end{table*}

As a result, work has been ongoing in establishing baselines and mapping the performance of monolingual models. Some of that work on high-resource languages is summarized in~\cite{scheible_gottbert_2020}. We highlight examples of published monolingual models for medium and low-resource languages along with English in Table~\ref{tab:1} with an emphasis on languages with resources similar to Icelandic in mC4. Generally, the model building approach is similar, although we note that in one case a multilingual model was used as a warm start~\cite{ralethe_adaptation_2020}.

We would also like to point out that for several languages of similar size to Icelandic (2.6B tokens) in mC4, there is no public monolingual BERT model available. These include Maltese (5.2B tokens), Kazakh (3.1B tokens), Georgian (2.5B tokens), Belarusian (2B tokens), Tajik (1.4B tokens), Kyrgyz (1B tokens), Somali (1.4B tokens), Sindhi (1.6B tokens), Armenian (2.4B tokens), and Luxembourgish (1B tokens). For others languages such as Macedonian (1.8B tokens), Malayalam (1.8B tokens), Mongolian (2.7B tokens), and Kannada (1.1B tokens) models exist online but to our best knowledge no publications exist that thoroughly document their performance on basic downstream tasks, such as PoS tagging and NER.

Another line of research has focused on how to make multilingual models more effective for low-resource languages. It has been shown that training on a larger corpus, such as filtered Common Crawl data, leads to significant improvements in downstream tasks, but increasing the number of languages beyond a certain point has a diluting effect that reduces overall performance~\cite{conneau_unsupervised_2020}. Others have shown that vocabulary extension of a multilingual model with continued pre-training on a monolingual corpus leads to improved performance and shorter training times than when starting from scratch~\cite{wang_extending_2020}.

The results on multilingual models indicate that language model performance is related to the amount of training data for the given language~\cite{xue_mt5_2021}, and studies on corpus quality indicate that the results are strongly related to the number of high quality sentences~\cite{kreutzer_quality_2021}. \newcite{kreutzer_quality_2021} have emphasized the importance of evaluating and auditing the corpora that are publicly available, since data in low-resource languages from multilingual datasets can be of low quality. They further emphasize the importance of developing high-quality evaluation datasets, since low-quality benchmarks might exaggerate model performance, making NLP for low-resource languages look further developed than it actually is.

\subsection{NLP for Icelandic}
Icelandic is a language from the West Germanic language family, with a rich morphology, where nouns, adjectives and verbs are highly inflected, and compounding is used actively to construct new words. The status of language data and resources for Icelandic is steadily improving, providing us with various datasets for evaluating our models, and benchmarks to measure against.

A good deal of work has been done on NLP for Icelandic that concerns these benchmarks. PoS tagging is implemented using a rule-based approach in the IceNLP toolkit~\cite{loftsson2007icenlp}, and using a Bi-LSTM model in the ABLTagger~\cite{steingrimsson-etal-2019-augmenting}. Constituency parsing has been implemented using a hand-crafted context-free grammar in the Greynir package~\cite{thorsteinsson-etal-2019-wide}, using finite-state transducers in IceParser~\cite{loftsson_iceparser_2007}, and using an mBERT model in~\cite{arnardottir2020neural}. NER for Icelandic has been implemented using a Bi-LSTM model and an ensemble tagger~\cite{ingolfsdottir_named_2020}.

\section{Training data}
The Icelandic datasets used for pre-training our models are listed in Table \ref{table:icebert_data}. They were split into validation, test and training sets and then tokenized~\cite{tokenizer}. We also do experiments on the Icelandic subset of the mC4 dataset (not shown in the table, see Section \ref{sec.3.2}), a dataset that is, in similar fashion to IC3, extracted from the Common Crawl but via a different method. %Having validation sets from each source proved to be particularly beneficial in monitoring performance by domain when training.

\begin{table}[!h]
\begin{center}
 \begin{tabularx}{\columnwidth}{X r r}
    \toprule
    \textbf{Dataset} & \textbf{Size }& \textbf{Tokens} \\
    \midrule
        IGC \emph{(editorial text)} & 8.2GB & 1,388M  \\
        IC3 \emph{(cleaned webcrawl)} & 4.9GB & 824M \\
        Student theses & 2.2GB & 367M \\
        Greynir News articles & 456MB & 76M \\
        Medical library & 33MB & 5.2M \\ 
        Open Icelandic e-books & 14MB & 2.6M \\
        Icelandic Sagas & 9MB & 1.7M \\\midrule
        \textbf{Total} & 15.8GB & 2,664M\\
        \bottomrule
  \end{tabularx}
    \caption[Icelandic texts used for training.]{Icelandic texts used for training.}% Sports news, which are highly repetitive and homogeneous, were removed from the IGC.}
\label{table:icebert_data}
\end{center}
\end{table}

The IGC~\citelanguageresource{risamalheild} is the most extensive collection of curated Icelandic text available. The IGC is mostly made up of news, legal documents and other copy-edited content and might, therefore, not accurately reflect the distribution of text from online sources. To supplement this dataset, several other sources were collected for pretraining the language models, as listed in Table \ref{table:icebert_data}. At the time of training the model, large collections of Icelandic literature were not available through legal means, but the recently updated IGC now includes some literary texts~\citelanguageresource{igcbooks}, which will be incorporated in future models. Social media and internet forum texts have now also been added to the updated IGC~\citelanguageresource{igcsocial}.

The IC3, our corpus of scraped and cleaned web texts (see next section for details), contains large amounts of text of many domains, topics, and styles at varying degrees of polish, and thus serves well as a complement to the IGC. In addition to the already mentioned data, academic texts found in student theses\footnote{\href{https://skemman.is}{\texttt{skemman.is}}} and data from the medical library of the University Hospital of Iceland\footnote{\href{https://www.hirsla.lsh.is}{\texttt{www.hirsla.lsh.is}}} were collected. The academic texts were passed through a filter reminiscent of the one used for the IC3 described in the next section, after an initial PDF text-extraction step. We also use texts scraped from Icelandic online news sites by the Greynir NLP engine\footnote{\href{https://greynir.is}{\texttt{greynir.is}}}.

\subsection{The Icelandic Common Crawl Corpus}
The Common Crawl Foundation is a non-profit organization that scrapes large semi-random subsets of the internet regularly and hosts timestamped and compressed dumps of the web online\footnote{ \href{https://commoncrawl.org/the-data/get-started/}{\texttt{commoncrawl.org/the-data/get-started/}}}. 
Each dump contains billions of web pages occupying hundreds of terabytes. Parsing these files directly requires storage and computing power not directly available to most and can come at a significant financial cost. The foundation also hosts indices of URIs and their locations within the large zipped dump files. While these indices are also large, their processing is feasible with a few terabytes of storage.

\subsubsection{Extracting Icelandic Common Crawl data}
The Common Crawl indices, which contain URI and byte offsets within the compressed dumps, are used to reduce the search space when looking for Icelandic texts. The Common Crawl Index Server has a public API\footnote{\href{https://index.commoncrawl.org}{\texttt{index.commoncrawl.org}}} where URIs can be queried based on attributes such as date, MIME-type and substring. Using the API eliminates the need to fetch the massive index files.

To extract Icelandic, the \texttt{.is} pattern is targeted to match the Icelandic top level domain (TLD), resulting in 63.5 million retrieved pages with URIs and byte locations within the compressed Common Crawl dumps. The computational efficiency of our method can be attributed to these steps. Given the predominant use of the \texttt{.is} TLD for Icelandic web content, we assume that other TLDs have a much lower proportion of Icelandic content. That said, a nontrivial amount of text in Icelandic is still likely to be found outside the \texttt{.is} domain and could be extracted by, e.g., parsing the whole Common Crawl, albeit at a much higher computational cost.

By targeting only the byte-offsets corresponding to the Icelandic TLD we extract candidate websites that have a high proportion of Icelandic content. In total, the compressed content is 687GiB on disk. All dumps since the start of the Common Crawl in 2008 until March 2020 were included. 

Plain text was extracted from the collected WARC (Web Archive format) files using jusText \cite{pomikalek2011removing}\footnote{We use the implementation at \texttt{https://github.com/miso-belica/jusText}.} to remove boilerplate content and HTML tags.
\subsubsection{Processing Common Crawl}

Once plain text had been extracted from the WARC files, Icelandic text was taken aside and duplicates removed. Since the \texttt{.is} TLD contains text in numerous languages, we use a fastText \cite{bojanowski2017enriching} model for extracting Icelandic text. Since the web is abundant with duplicate or near duplicate content, the data is first deduplicated at the document level and then at the inter-sentence level by sliding a three-line window over the text. If any three consecutive lines have appeared together previously, they are discarded. This latter step removes a fair amount of unwanted content, such as cookie notifications and thumbnail text. A summary of the filtering steps taken is shown in Table \ref{table:ic3desc}.

\begin{table}[!h]
\begin{center}
\begin{tabularx}{\columnwidth}{X r r}
\toprule \textbf{Filtering step} & \textbf{Size} & \textbf{\%} \\ \midrule
\texttt{.is} TLD & 687GB & 100\% \\
IS lang. filter and boilerpl. rem. & 29GB & 4.2\% \\
Dedup. document & 8.6GB & 1.3\% \\
Dedup. window & 4.9GB & 0.71\% \\
\bottomrule
\end{tabularx}
\caption[IC3: Filtering steps and retained data]{Filtering steps and retained data for IC3.}
\label{table:ic3desc}
\end{center}
\end{table}

\subsubsection{Comparison between IGC and IC3}

The two corpora, IC3 and IGC, are significantly different at the level of individual words. There are 1,155k unique tokens in the IC3 and 1,434k unique tokens in IGC, of which only 818k are shared.\footnote{If tokens with a count below 5 are not excluded there are 6.5M unique tokens in IGC and 6M unique tokens in IC3.} Almost one-third of the unique tokens (337k) in the IC3 are not present in IGC, and almost half of the IGC tokens (616k) are not present in IC3.

\subsection{The Icelandic part of mC4}\label{sec.3.2}
The Icelandic part of mC4 (mC4-is) contains 2.6B tokens (\textasciitilde 8GB on disk). The data was not filtered in the same way as the IC3 which is reflected by the masked token perplexity results shown in Table~\ref{table:icebert}; we hypothesize that further processing would be necessary to make use of it. By eyeing a random subset of the data, we see that a fair amount is badly machine-translated and some segments contain a lot of noise that is non-alphanumeric or otherwise not fluid text. If further processed, the results might very well be similar to that of IC3, but this analysis is left as future work.

\section{Training language models}

We train four different models following the RoBERTa-base architecture~\cite{roberta}, using 48 Nvidia 32GB V100 GPUs for approximately two days or 225k updates, with a batch size of \textasciitilde 955k tokens (2k sequences). We also train a single model using the RoBERTa-large architecture. This model became unstable in training after 37.5k steps with a batch size of \textasciitilde 4M tokens (8k sentences) and we did not make attempts to further improve it, but we fine-tune it in our experiments for comparison with the base models. All models use the same BPE-vocabulary, containing 50k tokens, constructed in the same way as the RoBERTa vocabulary. We train the models using four different data settings: all of the data available except mC4-is (IceBERT and IceBERT-large); the IC3 dataset (IceBERT-IC3), the IGC dataset (IceBERT-IGC); and mC4-is (IceBERT-mC4-is).

Furthermore, we evaluate performance using the multilingual XLMR-base \cite{conneau_unsupervised_2020} model as-is. We also experiment with continued pre-training on the IC3 corpus. Two models are trained: One for 100k steps with a batch size of 40k tokens and the other for 225k steps with a batch size of 80k tokens. The first model took one GPU-day in training and the other seven GPU-days. We do this to show what performance can be gained from leveraging a publicly available multilingual while minimizing computation cost as the seven day model uses about 8\% of the GPU hours used for training of the IceBERT-base models.

\begin{table}[h]
  \centering
\begin{tabularx}{\columnwidth}{X r r r r}
   \toprule
\textbf{Dataset} & \textbf{All*} & \textbf{IC3} & \textbf{IGC} & \textbf{mC4-is} \\ \midrule
IGC  & 3.64 & 4.49 & 3.51 & 16.50\\
IC3 & 4.40 & 4.15 & 5.67 & 17.84\\
Student theses & 4.12 & 4.95 & 5.47 & 15.40\\
Medical library & 5.26 & 5.87 & 7.26 & 20.49\\
Greynir News & 3.71 & 4.43 & 4.05 & 14.11\\
Icelandic Sagas & 7.53 & 7.09 & 12.49 & 52.00\\
Icelandic e-books & 9.04 & 8.75 & 10.37 & 36.48\\\bottomrule
\end{tabularx}
\caption[IceBERT masked token perplexity over development sets]{Masked token perplexity over development sets using models trained on different Icelandic datasets and combinations thereof. The All* model refers to IceBERT trained on all data except mC4-is.}
\label{table:icebert}
\end{table}

\section{Results}
We fine-tune and adapt the different IceBERT models for several classification and parsing tasks with state-of-the-art results after fine-tuning. Where F-scores are reported they are macro-averaged.

For PoS labelling and constituency parsing we extend the \texttt{fairseq} library; the resulting package \texttt{greynirseq} has been made available\footnote{See \href{https://github.com/mideind/greynirseq}{\texttt{github.com/mideind/greynirseq}} and the package \texttt{greynirseq} on PyPI}. We use the implementation in \texttt{fairseq} to evaluate performance on the Icelandic WinoGrande dataset.

For named entity recognition and grammatical error detection, we use the \texttt{transformers} library from Hugging Face \cite{huggingface}. For this purpose, we convert IceBERT to work with the library\footnote{The resulting model is available for use at \href{https://huggingface.co/mideind/IceBERT}{\texttt{huggingface.co/mideind/IceBERT}}.}.
\subsection{Part of Speech}

We fine-tune our models for PoS tagging using \texttt{greynirseq} on the MIM-GOLD \citelanguageresource{mimgold} dataset using ten-fold cross validation. The best performing models reach an accuracy of 98.4\%. We exclude the \emph{x} (not analyzed due to e.g. incorrect spelling) and \emph{e} (foreign) labels. In comparison, \cite{steingrimsson-etal-2019-augmenting} achieve 94.04\% accuracy. The results are shown in Table \ref{table:pos}.

In contrast to prior work on Icelandic PoS tagging, which universally approaches this as a multi-class classification task, we use a multi-label multi-class approach where we predict grammatical categories (gender, tense, etc.) independently instead of all together in one label. We adopt this approach to address a significant label scarcity problem in the training set and to allow for better generalization. See appendix \ref{appendix:pos} for a more comprehensive description.

We train the PoS models with a batch size of 32 sentences for 5 epochs. Peak learning rate is 5-e5 with approximately 0.2 epochs for warmup and a linear decay to zero. For the randomly initialized (no pretraining) model we do an additional longer run on the data since the model was clearly nowhere near convergence after 5 epochs, this was only done on a single split of data due to time constraints.

\begin{table}[!h]
\begin{center}
\begin{tabularx}{\columnwidth}{X r}
\toprule \textbf{Model} & \textbf{Accuracy} \\ \midrule
IceBERT & 98.33 $\pm$ 0.05 \\
IceBERT-large & 98.35  $\pm$ 0.06 \\
IceBERT-IGC & 98.27 $\pm$ 0.05 \\
IceBERT-IC3 & 98.30 $\pm$ 0.05 \\
IceBERT-mC4-is & 97.62 $\pm$ 0.10 \\
XLMR-base & 96.70 $\pm$ 0.15 \\
XLMR-base-IC3-1d & 97.20 $\pm$ 0.10 \\
XLMR-base-IC3-7d & 98.20 $\pm$ 0.07 \\
No pretraining (5 epochs) & 74.96 $\pm$ 0.54 \\
No pretraining (50 epochs) & 90.27 \\
\bottomrule
\end{tabularx}
\caption{Comparison of PoS-tagging performance for the models considered, including randomly initialized models.}
\label{table:pos}
\end{center}
\end{table}

All of the Icelandic models (except the one trained on mC4-is) show similar results of \textasciitilde98.3\% accuracy. An informal review of the errors that the best models make leads us to believe that this performance is about as good as it gets with this dataset and model architecture. The majority of errors can be classified as either being problems with the reference data or due to inherently ambiguous sentences. Problems with the reference data are either mislabeled examples or inconsistently applied rules, especially around proper nouns. Ambiguous sentences are mostly due to pronouns whose gender is unknowable without longer context. Most of the remaining errors are very difficult examples that require extensive world knowledge or complex co-reference resolution.

\subsection{Named Entity Recognition}

When fine-tuning IceBERT for named entity recognition (NER), it reaches state-of-the-art performance, showing a considerable improvement over the prior result of 85.79 macro F1-score in \citelanguageresource{ingolfsdottir_named_2020}. 
In fine-tuning we use a batch size of 16 sentences, peak learning rate of 2e-5 and chose the highest performing model on the validation set as measured across 10 epochs. The results over the test set for the different models each averaged over five seeds are shown in table \ref{table:ner}.

\begin{table*}[!h]
\begin{center}
\begin{tabularx}{\textwidth}{X r r r r}
\toprule \textbf{Model} & \textbf{F1} & \textbf{Prec.} & \textbf{Rec.} & \textbf{Acc.} \\ \midrule
IceBERT & 91.43 $\pm$ 0.23 & 91.60 $\pm$ 0.13 & 91.26 $\pm$ 0.36 & 98.66 $\pm$ 0.03 \\
IceBERT-large & 91.20 $\pm$ 0.36 & 90.79 $\pm$ 0.84 & 91.61 $\pm$ 0.41 & 98.58 $\pm$ 0.08 \\
IceBERT-IGC & 91.10 $\pm$ 0.25 & 91.15 $\pm$ 0.38 & 91.06 $\pm$ 0.20 & 98.59 $\pm$ 0.04 \\
IceBERT-IC3 & 91.29 $\pm$ 0.16 & 91.24 $\pm$ 0.24 & 91.35 $\pm$ 0.27 & 98.62 $\pm$ 0.02 \\
IceBERT-mC4-is & 89.57 $\pm$ 0.28 & 89.27 $\pm$ 0.44 & 89.87 $\pm$ 0.28 & 98.40 $\pm$ 0.06 \\
XLMR-base & 88.95 $\pm$ 0.60 & 88.66 $\pm$ 0.78 & 89.25 $\pm$ 0.61 & 98.41 $\pm$ 0.08 \\
XLMR-base-IC3-1d & 89.58 $\pm$ 0.30 & 89.84 $\pm$ 0.46 & 89.33 $\pm$ 0.26 & 98.39 $\pm$ 0.04 \\
XLMR-base-IC3-7d & 92.52 $\pm$ 0.40 & 92.31 $\pm$ 0.49 & 92.74 $\pm$ 0.41 & 98.83 $\pm$ 0.05\\
\bottomrule
\end{tabularx}
\caption{NER performance for models trained on different datasets, standard deviation over five seeds shown.}
\label{table:ner}
\end{center}
\end{table*}

The results are similar for all monolingual models and show that a lot of data or curated editorial corpora in pre-training are not necessary to achieve competitive NER performance. We would like to highlight that the XLMR-base multilingual model trained for 7-days on IC3 performs best on this task.

\subsection{Constituency parsing}
We implement a simplified version of the CKY-style chart parser described by \cite{kitaev-etal-2019-multilingual}\footnote{Implemented in the \texttt{greynirseq} repository.}. We did not implement position-factored attention nor incorporate any extra word features, such as PoS or character information, since our goal is primarily to measure the knowledge captured by the model. We leave such experiments for future work.

The dataset we use is GreynirCorpus~\citelanguageresource{greynircorpus}, a constituency annotated version of the aforementioned Greynir News dataset, whose test and development sets are human-annotated. Its annotation scheme comes from the Greynir rule-based parser \cite{thorsteinsson-etal-2019-wide} and shares many similarities with the Penn Treebank-derived \citelanguageresource{marcus-etal-1993-building} schemas and their corresponding annotation guidelines.

A generalized version of the GreynirCorpus test set was created for a fairer comparison with previous parsers for Icelandic, namely the Greynir parser, IceParser \cite{loftsson-rognvaldsson-2007-iceparser} — a shallow parser for Icelandic, and a variant of the Berkeley Neural Parser \cite{arnardottir2020neural} which comprises a multilingual BERT fine-tuned on the Icelandic Parsed Historical Corpus (IcePaHC) \cite{rognvaldsson-etal-2012-icelandic}.

We split the development subset of the GreynirCorpus into ad-hoc train and validation splits and train on the respective portion. Results from testing on the generalized benchmark mentioned above are shown in Table \ref{table:parsing}. Somewhat surprisingly, the large model does not show best performance on the task, we believe that this would change if the model is trained to convergence or better hyperparameter tuning. We note that differences between models are only slight as all the models are within 2 percentage points from each other. For extra comparison, a randomly initialized model did not surpass 70 F1-score.

%  For a more detailed description see the appendix.
\begin{table*}[h]
\begin{center}
\begin{tabularx}{\textwidth}{X r r r r}
\toprule \textbf{Model} & \textbf{F1} & \textbf{Prec.} & \textbf{Rec.} \\ \midrule
% \hline \textbf{Model} & \textbf{F-score} & \textbf{Prec} & \textbf{Rec} & \textbf{Cross} \\ \hline
IceBERT & 90.02 $\pm$ 0.12 & 87.93 $\pm$ 0.16 & 92.20 $\pm$ 0.07 \\ % & 0.55 \\
IceBERT-large & 89.79 $\pm$ 0.13 & 87.71 $\pm$ 0.28 & 91.98 $\pm$ 0.14 \\
IceBERT-IGC & 89.66 $\pm$ 0.12 & 87.20 $\pm$ 0.13 & 92.25 $\pm$ 0.15 \\
IceBERT-IC3 & 89.37 $\pm$ 0.14 & 86.73 $\pm$ 0.32 & 92.18 $\pm$ 0.17 \\
IceBERT-mC4-is & 88.60 $\pm$ 0.16 & 86.08 $\pm$ 0.13 & 91.27 $\pm$ 0.40 \\
XLMR-base & 88.16 $\pm$ 0.27 & 86.16 $\pm$ 0.33 & 90.26 $\pm$ 0.24 \\
XLMR-base-IC3-1d & 88.67 $\pm$ 0.16 & 86.74 $\pm$ 0.10 & 90.75 $\pm$ 0.24 \\
XLMR-base-IC3-7d & 89.01 $\pm$ 0.09 & 86.95 $\pm$ 0.12 & 91.16 $\pm$ 0.09 \\
\bottomrule
\end{tabularx}
\caption{EVALB performance on the generalized form of the GreynirCorpus test set, mean and standard deviation over five seeds shown.}
\label{table:parsing}
\end{center}
\end{table*}

\subsection{Grammatical error detection}
We fine-tune IceBERT for grammatical error detection (GED). These are the first machine learning models trained for GED in Icelandic, and make use of the Icelandic Error Corpus (IceEC) \citelanguageresource{iceec1.1} which contains 58k labeled sentences.

We train models for both binary and multi-class token-level classification. For the multi-class GED classifier, we exclude ambiguous sentences from the IceEC where a single token has multiple error labels; this removes 3.5k sentences out of the 23k sentences with error labels. The problem could be solved as a multi-label one, but we leave that approach for future work. While the dataset is fine-grained and contains a variety of objective and subjective labels (e.g.\ stylistic), in this study we limit our evaluation to the five high-level categories \emph{coherence, grammar, orthography, style and vocabulary}.

\begin{table*}[!h]
\begin{center}
\begin{tabularx}{\textwidth}{X r r r r}
\toprule \textbf{Model} & \textbf{F1} & \textbf{Prec.} & \textbf{Rec.} & \textbf{Acc.} \\ \midrule
IceBERT & 70.11 $\pm$ 0.91 & 92.30 $\pm$ 0.17 & 56.53 $\pm$ 1.12 & 96.99 $\pm$ 0.07 \\ % 69.98 & 90.97 & 56.86 & 96.95 \\
IceBERT-large & 89.12 $\pm$ 1.31 & 93.55 $\pm$ 0.67 & 85.10 $\pm$ 1.87 & 98.70 $\pm$ 0.15 \\% & 88.04 & 92.48 & 84.01 & 98.58 \\
IceBERT-IGC & 71.33 $\pm$ 1.65 & 92.14 $\pm$ 0.42 & 58.22 $\pm$ 2.08 & 97.08 $\pm$ 0.13 \\%& 71.71 & 92.77 & 58.44 & 97.12 \\
IceBERT-IC3 & 70.95 $\pm$ 1.10 & 91.97 $\pm$ 0.62 & 57.77 $\pm$ 1.41 & 97.05 $\pm$ 0.09
\\%& 68.90 & 90.97 & 56.86 & 96.95 \\
IceBERT-mC4-is & 57.72 $\pm$ 1.07 & 90.98 $\pm$ 0.58 & 42.28 $\pm$ 1.20 & 96.13 $\pm$ 0.06 \\%& 59.87 & 87.68 & 45.46 & 96.20 \\
XLMR-base & 64.52 $\pm$ 1.82 & 88.11 $\pm$ 0.72 & 50.93 $\pm$ 2.21 & 96.75 $\pm$ 0.12
\\%& 64.65 & 84.57 & 54.32 & 96.67 \\
XLMR-base-IC3-1d & 62.73 $\pm$ 3.05 & 86.62 $\pm$ 1.55 & 49.22 $\pm$ 3.38 & 96.61 $\pm$ 0.21 \\%& 44.74 & 85.06 & 30.35 & 96.64 \\
XLMR-base-IC3-7d & 75.87 $\pm$ 1.02 & 91.64 $\pm$ 0.23 & 64.74 $\pm$ 1.43 & 97.61 $\pm$ 0.08 \\
\bottomrule
\end{tabularx}
\caption{Binary token classification performance measured using the Icelandic Error Corpus evaluation dataset, standard deviation over five seeds shown.}
\label{table:ged-token-binary}
\end{center}
\end{table*}

For fine-tuning, we use a batch size of 16, learning rate of 2e-5 and train five times for five epochs with different seeds. The results for binary classification are shown in Table \ref{table:ged-token-binary} and by category for the top 5 models in the multi-class task in Table \ref{table:ged-token-category}. For the three lowest performing models IceBERT-mC4-is had a total accuracy of 51.69 $\pm$ 2.60, XLMR had a total accuracy of 56.26 $\pm$ 4.22, and XLMR-IC3-1d had a total accuracy of 41.24 $\pm$ 0.60.

Based on the experiments it is clear that out of the IceBERT base-models, the model trained on the IGC dataset containing editorial text is best suited for fine-tuning for GED. IceBERT-large is the clear winner with an F1 score of 89.12 $\pm$ 1.31. The 1-day XLMR model does not do as well as the IceBERT models and the mC4-is model is lagging behind, further highlighting that models trained on the dataset might benefit from further cleanup. Interestingly, we see that the 7-day XLMR multilingual model outperforms the other models besides IceBERT-large. 

\begin{table*}[!h]
\begin{center}

\begin{tabularx}{\textwidth}{X r r r r r}
\toprule  \textbf{Category} & \textbf{IB-base} & \textbf{IB-large} & \textbf{IB-IGC} & \textbf{IB-IC3} &  \textbf{XLMR-IC3-7d}\\ \midrule
\textbf{Coherence}  & 5.90 $\pm$ 3.15 & 41.58 $\pm$ 21.83 & 6.89 $\pm$ 3.68 & 4.94 $\pm$ 3.23 & 2.40 $\pm$ 3.00 \\ % & 8.23 & 55.32 & 7.73 & 7.24 & 0 & 0 \\
\textbf{Grammar} & 55.05 $\pm$ 2.64  & 72.09 $\pm$ 14.25 & 54.71 $\pm$ 4.77 & 50.71 $\pm$ 4.93 & 62.83 $\pm$ 3.69 \\ % 54.41 & 83.31 & 59.25 & 55.46 & 37.22 & 23.96 \\
\textbf{Orthography} & 80.39 $\pm$ 2.12  & 89.66 $\pm$ 4.88 & 81.00 $\pm$ 2.30 & 79.79 $\pm$ 2.29 & 84.47 $\pm$ 2.19 \\ %& 81.20 & 93.42 & 83.47 & 74.97 & 73.49 & 79.08 \\
%Other & 0 & 0 & 0 & 0 \\
\textbf{Style} & 36.07 $\pm$ 10.42  & 67.39 $\pm$ 16.30 &  37.19 $\pm$ 9.66 & 35.50 $\pm$ 9.66 & 47.44 $\pm$ 9.02 \\ %& 41.42 & 80.53 & 46.14 & 38.66 & 26.15 & 24.44 \\
\textbf{Vocabulary} & 18.47 $\pm$ 2.74  & 50.30 $\pm$ 21.45 & 17.94 $\pm$ 4.48 & 16.85 $\pm$ 4.45 & 17.33 $\pm$ 4.56\\ %& 20.38 & 64.91 & 20.47 & 18.38 & 11.39 & 9.24 \\ 
\midrule
All & 62.83 $\pm$ 3.75  & 79.16 $\pm$ 9.39 & 63.45 $\pm$ 3.96 & 61.84 $\pm$ 4.06 & 69.42 $\pm$ 3.58 \\ %& 64.42 & 86.50 & 67.37 & 64.35 & 53.88 & 57.79 \\
\bottomrule
\end{tabularx}
\caption{Token classification F1-scores measured using the Icelandic Error Corpus evaluation dataset for the top-5 highest scoring models. }
\label{table:ged-token-category}
\end{center}
\end{table*}

\subsection{Icelandic WinoGrande}
The WinoGrande dataset \citelanguageresource{Sakaguchi2020WINOGRANDEAA}, used for evaluating commonsense reasoning capabilities of neural language models, is inspired by the original WinoGrad dataset~\cite{levesque2012winograd}, but its problems are designed to minimize biases which the models may rely on when solving them. The dataset consists of sentences that include two nouns and an ambiguous pronoun which grammatically can refer to either of those noun phrases. The task is to decide which noun makes more semantic sense, given the information in the sentence.

We systematically go through the WinoGrande test set (1767 examples) and manually translate and adapt sentences to work in Icelandic. While the English WinoGrande problems are not always constructed as pairs, in our adaptation, we create sentence pairs where it is feasible. We also found some of the examples to be specific to culture, subjective, or otherwise inapplicable for translation. Those examples were either adjusted or skipped. The result is a dataset of 1095 examples. The size of the Icelandic dataset is closest in size to the small variant of the English dataset (640 examples).

\begin{table}[!h]
\begin{center}
\begin{tabularx}{\columnwidth}{X r}
\toprule \textbf{Model} & \textbf{Accuracy} \\ \midrule
IceBERT & 54.6 $\pm$ 2.1 \% \\
IceBERT-large & 57.1  $\pm$ 3.7 \% \\
IceBERT-IGC & 53.8 $\pm$ 2.3 \% \\
IceBERT-IC3 & 53.8 $\pm$ 2.4 \% \\
IceBERT-mC4-is & 51.4 $\pm$ 2.1 \% \\
XLMR-base & 52.4 $\pm$ 1.7 \% \\
XLMR-base-IC3-1d & 51.2 $\pm$ 2.0 \% \\
XLMR-base-IC3-7d & 53.4 $\pm$ 1.9 \% \\
\bottomrule
\end{tabularx}
\caption{Accuracy on the Icelandic WinoGrande dataset. Results are averaged over five-fold cross-validation and standard deviation is reported.}
\label{table:winogrande}
\end{center}
\end{table}

The results over five-fold cross-validation can be seen in Table \ref{table:winogrande}. The large model outperforms the other variants while models trained on IGC and/or IC3 outperform the model trained on mC4-is. The XLMR-base and XLMR-base-IC3-1d perform similar to the model trained on mC4-is but the XLMR-base-IC3-7d performs similar to the model trained only on IC3 or IGC.

\section{Conclusion}
We have successfully trained baseline neural language models for Icelandic that perform well on existing benchmarks, in particular NER, PoS and constituency parsing. We also present the Icelandic WinoGrande dataset and show that it is challenging for the models we evaluate.

Furthermore, to our surprise, we show that extracting data from online sources is sufficient to train models which show performance that is competitive with those trained on curated/editorial corpora. We stress that proper filtering and cleanup of crawled data is necessary, as demonstrated in the difference between training models on IC3 and mC4-is.

For some downstream tasks, we observe that multilingual language models~\cite{conneau_unsupervised_2020} are getting quite close to the performance of models trained on a monolingual corpus but for other tasks we still observe a significant difference. Interestingly, using such models as a warm start for some tasks can even lead to state-of-the-art performance. 

We conclude that by using text extracted from the Common Crawl corpus and multilingual models as a warm start, well-performing language models are becoming more feasible to build for low to medium-resource languages. We still hold that curated corpora are beneficial in certain applications, but the added value for downstream tasks is small if a sufficiently large crawled corpus is available. 

\section{Acknowledgements}
We thank Prof. Dr.-Ing. Morris Riedel and his team for providing access to the DEEP super-computer at Forschungszentrum Jülich. We also thank the Icelandic Language Technology Program~\cite{nikulasdottir_language_2020}. It has enabled the authors to focus on work in Icelandic NLP.

%\appendix
\begin{appendices}

\section{Part-of-speech tagging}
\label{appendix:pos}
We use the MIM-GOLD dataset \citelanguageresource{mimgold} to train a PoS tagger. If one approaches this task as a multiclass problem (each word gets exactly one label) there is a significant label scarcity problem. There are approximately 600 legal labels in the MIM-GOLD schema, but only 559 of them appear in the training set, and 43 of them appear less than 10 times. The 10th, 25th, 50th and 75th percentile tags occur 15, 52, 208 and 728 times, respectively.

To address this problem we decompose the labels. Instead of considering a label to be a single unit, e.g. \emph{noun-masculine-singular-nominative-article}, we consider it to be composed of several categories, e.g. \emph{lexical class}, \emph{gender}, \emph{number}, \emph{case} and \emph{article-clitic}, each of which can have several values. We also observe that some categories are shared between lexical class, e.g. nouns and adjectives both have gender, number and case and typically share values when they co-refer. Our model therefore outputs for each word a lexical class and a value for every grammatical category or morphological feature, but we ignore those that are not applicable to the lexical class, e.g. for nouns we mask out loss for the \emph{tense} category during training and output no \emph{tense} label during inference.

Since the number of labels within each category is small (the largest category has 6 possible labels), each label has been seen many times during training, even though some combinations of labels never occur in the training set (such as \emph{verb past subjunctive 2person plural middle-voice}). This allows the model to generalize and predict these unseen combinations, some of which actually occur in the test set.

\end{appendices}

% \nocite{*}
\section{Bibliographical References}\label{reference}
%\label{main:ref}

\bibliographystyle{lrec2022-bib}
\bibliography{lrec2022-example}

\section{Language Resource References}
\label{lr:ref}
\bibliographystylelanguageresource{lrec2022-bib}
\bibliographylanguageresource{languageresource}
\end{document}